\documentclass[journal,twocolumn,10pt]{IEEEtran}\usepackage{algorithm}   
\usepackage{algpseudocode}%
\usepackage{amsmath,amsfonts}

\usepackage[caption=false,font=normalsize,labelfont=sf,textfont=sf]{subfig}
\usepackage{textcomp}
\usepackage{stfloats}
\usepackage{verbatim}
\usepackage{graphicx}
\usepackage{multirow}
\usepackage{cite}
\usepackage{booktabs}   
\usepackage{xcolor}    
\usepackage{rotating}    
\usepackage{array}    
\usepackage{threeparttable}
\usepackage[table]{xcolor}
\usepackage{amsmath}
\usepackage{xcolor,soul,framed} 
\usepackage{amssymb}
\usepackage{tabularx}  
\usepackage{multirow} 
\usepackage{graphicx}  
\usepackage{booktabs}  
\usepackage{xcolor}    
\usepackage{makecell}   
\colorlet{shadecolor}{yellow}
\graphicspath{{../pdf/}{../jpeg/}}
\DeclareGraphicsExtensions{.pdf,.jpeg,.png}

\usepackage{array}
\usepackage{mdwmath}
\usepackage{mdwtab}
\usepackage{eqparbox}
\usepackage{url}
\usepackage{amssymb}
\usepackage{multirow}
\usepackage{booktabs}
\usepackage{subcaption}  
\usepackage{xcolor} 
\usepackage[numbers]{natbib} 
\begin{document}
\bstctlcite{IEEEexample:BSTcontrol}
\title{DCG ReID: Disentangling Collaboration and Guidance Fusion Representations for Multi-modal Vehicle Re-Identification}

\author{Aihua Zheng, Ya Gao, Shihao Li, Chenglong Li*, Jin Tang, ~\IEEEmembership{\textit{Senior Member, IEEE}}
\thanks{This research is supported in part by the National Natural Science Foundation of China under Grants 62372003 and 62376004, and the Natural Science Foundation of Anhui Province under Grants 2308085Y40. (*The corresponding author is Chenglong Li.)}
\thanks{C. Li, A. Zheng, Y. Gao and S. Li are with the Information Materials and Intelligent Sensing Laboratory of Anhui Province, Anhui Provincial Key Laboratory of Security Artificial Intelligence, School of Artificial Intelligence,
Anhui University, Hefei, 230601, China (e-mail: lcl1314@foxmail.com; ahzheng214@foxmail.com; gaoya615@foxmail.com; shli0603@foxmail.com).}
\thanks{J. Tang is with Anhui Provincial Key Laboratory of Multimodal Cognitive Computation, School of Computer Science and Technology, Anhui University, Hefei, 230601, China (e-mail: ahu\_tj@163.com).}}

\markboth{Journal of \LaTeX\ Class Files,~Vol.~14, No.~8, August~2021}%
{Shell \MakeLowercase{\textit{et al.}}: A Sample Article Using IEEEtran.cls for IEEE Journals}


\maketitle

\begin{abstract}
Multi-modal vehicle Re-Identification (ReID) aims to leverage complementary information from RGB, Near Infrared (NIR), and Thermal Infrared (TIR) modalities to retrieve the same vehicle.
The challenges of multi-modal vehicle Re-Identification (ReID) arise from the uncertainty of modality quality distribution induced by the inherent discrepancies across multiple modalities, which results in distinct yet conflicting fusion requirements for data with balanced and unbalanced quality distributions. 
Existing methods handle all multi-modal data within a single fusion model, which causes the model to overlook the different needs of the two types of data, making it difficult to decouple the conflict between intra-class consistency and inter-modal heterogeneity. 
To this end, we propose to Disentangle Collaboration and
Guidance Fusion Representations for Multi-modal Vehicle ReID (DCG-ReID). 
Specifically, to disentangle modal data with heterogeneous quality distributions while avoiding mutual interference, we first propose the Dynamic Confidence-based Disentangling Weighting (DCDW) mechanism, which realizes the disentangled fusion framework by dynamically reweighting the contributions of the three modalities via modal confidence derived from multi-modal interaction.
%
%
Building on the DCDW, we further develop two scenario-specific fusion strategies to tackle distinct quality distribution characteristics:
%
(1) for balanced multi-modal quality distributions, we introduce the Collaboration Fusion Module (CFM), which mines and extracts consensus features in a pairwise manner to capture shared discriminative information, in turn effectively enhances intra-class consistency; 
(2) for those with unbalanced quality distributions, we propose a Guidance Fusion Module (GFM), which implements differential amplification of modal discriminative disparities to reinforce the advantages of the dominant modality while guiding auxiliary modalities to mine complementary discriminative information, thereby effectively mitigating inter-modal divergence and substantially boosting multi-modal joint decision-making performance. 
%
The extensive experiments on three multi-modal vehicle ReID benchmarks WMVeID863, MSVR310 and RGBNT100 validate the effectiveness of our method. 
The code will be released upon acceptance.
\end{abstract}


\section{Introduction}
\IEEEPARstart{V}{ehicle} ReID task involves retrieving the same vehicle targets from non-overlapping camera views. 
Over the past decade, the recognition accuracy of ReID models has been continuously improving. Traditional vehicle ReID tasks are single-modal tasks, which use visible light image (RGB) for retrieval.
Although the rich color features of the RGB have significant discriminative advantages~\citep{Liu2016Deep, Guo2018Learning, Lou2019VERI-Wild:, Tang2019CityFlow:, Yu2024TF-CLIP:, tcsvt5, tcsvt6, tcsvt7, tcsvt8}, single modality vehicle ReID cannot handle complex scenarios such as night, insufficient light, and vehicle glare. 
The unique ability of TIR to perceive thermal radiation alleviates the problem of insufficient light, and the advantages of NIR in occlusion, flare-contamination and low-light environments also improve the performance of multi-modal vehicle ReID.
Therefore, the introduction of TIR~\cite{Facenet} and NIR~\cite{HAMNET} modalities when RGB lacks key discriminatory information has driven substantial progress in ReID. Zheng \textit{et al.}~\cite{CCNET310} and Li \textit{et al.}~\cite{100} first propose the multi-modal vehicle ReID research tasks and establish high-quality benchmarks. And there are researches utilizes multi-modal data to improve ReID model performance ranging from fine-grained alignment~\cite{Editor, tcsvt4, top} to coarse-grained interactive fusion~\cite{Demo, promptMA,ICPL-ReID,mambapro,IDEA:}.
\begin{figure}[t]
\centering
\includegraphics[width=0.48\textwidth, height=4.0in, keepaspectratio]{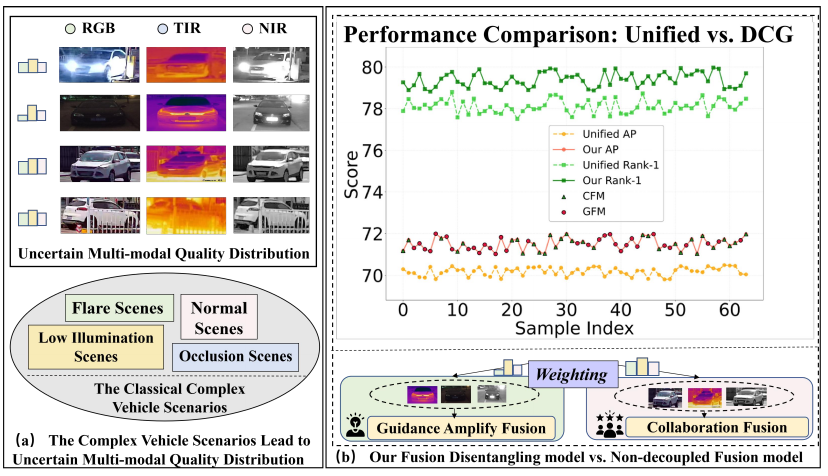}\label{fig_first_case}\\
\caption{
(a) Inherent key challenges in multi-modal vehicle ReID lead to unbalanced multi-modal quality distribution. 
%
%
(b) The heatmap of AP and Rank-1 gains on samples between our fusion decoupling method and the unified fusion model (i.e., it does not distinguish between the quality distributions of multi-modal samples); The basic approach of our DCG based on fusion decoupling.
}
\label{fig_1}
\end{figure}
Among these approaches, the gains brought by multi-modal can be further enhanced by incorporating multi-modal fusion.
Zheng \textit{et al.}~\cite{Facenet} utilize the introduced flare-immune TIR modality and fusing it with the other two modalities,  which enhances the recognition performance under the challenge of nighttime flares.
Yu \textit{et al.}~\cite{fusion1} propose windows from local to global which implemented to select important tokens, followed by token fusion and feature aggregation to extract key information of specific modalities while alleviating cross-modal differences. This further reduces cross-modal heterogeneity and enhances feature discriminative ability.

The introduction of multi-spectral technology not only brings new improvements to the recognition accuracy of models in ReID tasks but also poses new challenges~\cite{100, CCNET310, coen} of multi-modal alignment and fusion arising from modality discrepancies. As shown in Fig~\ref{fig_1} (a), the challenges arise from dynamically changing real-world scenarios (e.g., low illumination and flare interference), which lead to uncertain modality quality distributions of multi-modal information. 
Accompanying this is that the distribution of multi-modal data appears either balanced or unbalanced.
%
Existing methods attempt to alleviate these challenges through a unified fusion model, such as multi-modal complementary alignment and multi-modal dynamic fusion. For example, Wang \textit{et al.}~\cite{Demo} use the Mixture-of-Experts (MoE) fusion method to fuse single-modal, dual-modal, and tri-modal information. Zheng \textit{et al.}~\cite{coen} dynamically select the dominant modality to guide the learning of auxiliary modalities within a single fusion framework~\cite{Facenet}, which has also significantly improved the ReID performance on WMVeID863~\cite{Facenet} dataset with unbalanced quality. 
\begin{figure}[t]
\centering
\includegraphics[width=0.48\textwidth, height=4.0in, keepaspectratio]{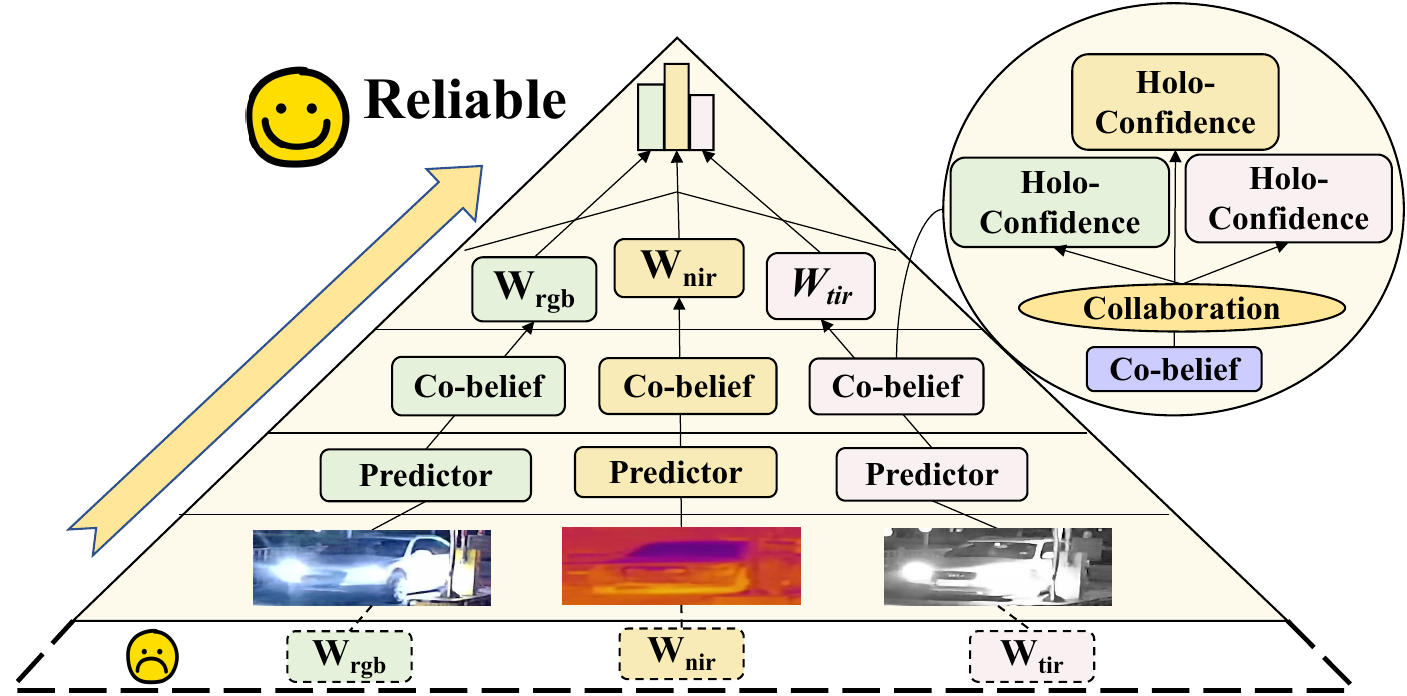}\label{fig_first_case}\\
\caption{Our reliable multi-modal dynamic weighting vs. single-modal weighting.}
\label{fig_2}
\end{figure}

However, as shown in Fig.~\ref{fig_1} (b), which compares the ReID evaluation (AP and Rank-1~\cite{Rank-1}) of each sample between the single fusion model and our fusion-decoupling model, the color heatmap indicates the R1 gain, and the height difference represents the sample AP gain. It is clearly found that for multi-modal fusion decisions with uncertain quality distributions, it is more appropriate to first perform fusion decoupling and then conduct targeted fusion according to the characteristics of the quality distribution.
The robustness of the decision of the traditional dynamic fusion is based on weight adjustment. When the quality of a certain modality drops sharply (e.g., RGB completely fails due to flare), the error-correcting capability of weight adjustment is limited, which easily leads to fusion results being dominated by noise. 
%
The core issue with unified fusion frameworks is that they overlook the inherent conflict between modalities with different quality distributions: when the model excessively focuses on dominant modal data, it will overfit to a specific modality among balanced-distribution modalities, thereby exacerbating the differences between modalities with balanced quality distributions; when the model places more emphasis on balanced-distribution modal data, it may neglect the advantageous features of the dominant modality, leading to failure in fully utilizing multi-modal information and, conversely, reducing intra-class consistency without leveraging the advantages of multi-modal.
%
Thus, accurately evaluating the dynamic characteristics of modality data quality has become the first key to breaking through fusion conflicts and further facilitating the full utilization of multi-modal information for final ReID decision-making.
%

%
Inspired by trustworthy fusion applications derived from trustworthy learning~\cite{predict}, relevant studies~\citep{Does, Xue2023Dynamic} on multi-modal trustworthy learning have shown that multi-modal learning can sometimes underperform compared with partial-modal or even single-modal learning, especially when there is an imbalance in modality quality. In corresponding downstream tasks, Qu \textit{et al.}~\cite{ijcai} have enhanced the performance of multi-modal fusion by leveraging trustworthy weight reallocation. 
As shown in the bottom of Fig.~\ref{fig_1}(b) and Fig.~\ref{fig_2}, we utilize interactive multi-modal weights to dynamically and ingeniously split multi-modal data into targeted fusion strategies by comprehensively considering the contributions of the three modalities. 
Based on this confidence-based decoupled fusion mechanism, we design two fusion strategies for the two types of data with different quality distributions, thereby fully exploiting multi-modal data while avoiding fusion conflicts. Meanwhile, the single-peak uncertainty of unimodal data raises a problem: the method of obtaining unimodal weights based on the modality's own confidence lacks comprehensive consideration for our multi-modal joint decision-making task. %
After obtaining the modal weight distribution in this way, we correspondingly decouple the two fusion frameworks to make targeted decisions respectively. 
Specifically, under the missing modality experimental setting for multi-modal ReID, we found that the fusion performance of some two-modal combinations outperforms that of three-modal fusion. Thus, we leverage such multi-modal weights to discard low-quality modal samples, avoiding the interference caused by inferior modal data. In the case of unbalanced multi-modal quality distribution, we amplify the advantages of the dominant modality and guide the learning of auxiliary modalities, thereby enhancing the reliability of multi-modal joint decision-making.

Specially, our Confidence-based Disentangling Collaboration and Guidance Fusion Representations for Multi-modal Vehicle ReID named DCG leverages Confidence-based Disentangling Weighting (DCDW) to differentiate multi-modal vehicle data with balanced and unbalanced quality distributions to avoid fusion conflicts. Furthermore, we design two targeted fusion measures for them to better utilize multi-modal data. Firstly, as shown in Fig.~\ref{fig_2}, DCDW calculates each modality’s mono-confidence and holo-confidence via multi-modal interactive evaluation. And integrate them namely co-belief to determine final weights, balancing modal contributions to cleverly identify the fusion conflicts and optimizing the targeted fusion.
Secondly, as shown in Fig.~\ref{fig_1}(b), based on the weights for each modality, we design two feature interaction fusion methods tailored to the different modality data quality distributions.
For balanced distribution, we leverage Collaboration Fusion Module (CFM) to employ a dropout collaborative learning approach which is based on confidence-based weights via a bidirectional enhancement attention mechanism. It maximizes consensus information extraction to expand the intra-class consistency by conducting pairwise mining between modalities.
Meanwhile, for unbalanced distributions, we design a Guidance Fusion Module (GFM) that leverages differential amplification with learnable factors. It amplifies unique information of the dominant modality while guiding auxiliary ones, thus reducing inter-modal heterogeneity through enhanced interactive learning.

In summary, we utilize multi-modal interactive confidence for decoupled fusion. While avoiding fusion conflicts caused by different data quality distributions, we design fusion strategies targeting two types of multi-modal data with balanced and unbalanced distribution. On the basis of resolving fusion conflicts, we fully leverage the positive gains brought by multi-modal data. The main contributions of our work are as follows:
\begin{itemize} 
\item To avoid mutual interference in fusion decision-making for uncertain quality multi-modal data distributions, we propose a Dynamic Confidence-Based Disentangling Weighting (DCDW) to decouple fusion based on adaptive confidence-based modal weights. 
\item We propose a Collaboration Fusion Module (CFM) to process modal features with balanced multi-modal quality distributions, mining shared information through pairwise inter-modal interactions and leveraging complementary shared information to expand intra-class consistency.
\item We propose a Guidance Fusion Module (GFM) for scenarios with unbalanced multi-modal quality distributions. On the basis of amplifying the advantageous parts of the dominant modality, it guides the learning of auxiliary modalities, while reducing inter-modal differences and enhancing the effectiveness of multi-modal interactive mining.
\item Extensive experiments on three multi-modal vehicle ReID datasets WMVeID863~\cite{Facenet}, MSVR310~\cite{CCNET310} and RGBNT100~\cite{100} validate the effectiveness of our method. 
\end{itemize}

\section{RELATED WORK}
\subsection{Multi-modal Vehicle ReID}
Multi-modal vehicle ReID uses data from diverse spectra, including RGB, NIR and TIR, to boost ReID task performance. Multi-modal vehicle ReID tasks often face challenges such as low light, occlusion, and flare contamination. Unlike RGB images, NIR can capture clear vehicle details such as license plate edges and headlight shapes even in dim environments, and TIR is immune to complete darkness and zero-visibility weather. As NIR and TIR have unique imaging features offering new solutions, Li \textit{et al.} propose RGBN300~\cite{HAMNET} dataset and construct a new baseline method HAMNet to fuse spectra features to enhance ReID robustness. Zheng \textit{et al.}~\cite{CCNET310} introduce a cross-directional network CCNet with a novel cross-directional center loss to simultaneously overcome the problems of cross-modality discrepancy and propose a benchmark dataset MSVR310. Wang \textit{et al.}~\cite{HTT} develop HTT which improves the representation ability of multi-modal by fine-tuning models on unseen test data using self-supervised loss for better generalization. To reduce background interference and reduce modality gaps, Zhang \textit{et al.}~\cite{Editor} propose a fine-coarse method which uses a spatial-frequency token selection module to filter discriminative tokens and aggregate token features named EDITOR, improving the feature discrimination with background suppression. Due to the immunity advantage of the TIR modality to vehicle flares, Zheng \textit{et al.}~\cite{Facenet} create the first low-quality multi-spectral vehicle ReID flare-affected dataset using TIR which remains robust under flare conditions, to enhance discriminative information. Zheng \textit{et al.}~\cite{coen} propose CoEN accounting for the impact of low quality, which is applicable to complex environments for selecting the dominant modality to collaboratively enhance other modalities. As the complex and intricate environments faced by vehicle ReID tasks, Wang \textit{et al.}~\cite{Demo} combine dynamic weighting with multi-expert fusion to handle uncertain multi-modal quality distributions. Yu \textit{et al.}~\cite{tcsvt4} design an Attention-Fourier Token Sparsification module to adaptively sparsify and connect tokens of multi-spectral images in the attention domain and Fourier domain, while putting forward an Information Unification Constraint learning strategy, thereby preserving spectrum-specific crucial information, aligning multi-spectral information. Jiao \textit{et al.}~\cite{tcsvt2} design a Pose-adaptive and Saliency Attention Network to address the challenges of feature deformation and invisibility of salient attributes caused by vehicle pose changes in aerial photography scenarios. In the unsupervised domain, Qiu \textit{et al.}~\cite{tcsvt3} for the first time exploited the feature distribution characteristics of vehicles within a single camera to optimize pseudo-label generation for unsupervised Re-ID. They further accurately leveraged the value of pseudo-labels through focal contrastive learning, addressing the core contradiction in unsupervised vehicle Re-ID of low reliance on annotations yet high pseudo-label noise and weak model discriminability, and providing an efficient solution for cross-camera vehicle recognition in real-world scenarios. While these methods have improved the recognition accuracy of multi-modal vehicle ReID models from different perspectives, they use a unified fusion model, risking conflict between intra-class consistency and inter-modal heterogeneity when modality quality distributions are different. In particular, performing full tri-modal fusion in all situations can be sub-optimal. In contrast, our proposed DCG utilize fusion decoupling dynamically adjusts the fusion strategy according to confidence-based modal weights. Moreover, on the basis of avoiding fusion conflicts, our method designs two fusion approaches respectively targeting uniform and non-uniform multi-modal quality distributions.
\begin{figure*}[t]
    \centering
    \includegraphics[width=1.0\linewidth]{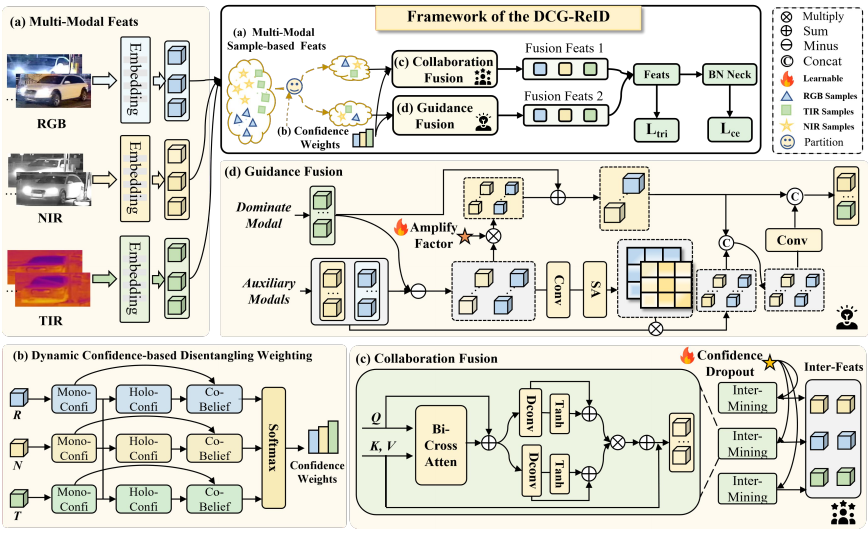}
    \caption{The framework of our DCG. Firstly, the three-modal features (RGB, NIR, TIR) are extracted by the shared CLIP visual encoder. They then enter our designed Dynamic Confidence-based Disentangling Weighting (DCDW) to obtain dynamic confidence-based weights. According to the weight distribution, we introduce two fusion strategies, namely the Collaboration Fusion Module (CFM) and the Guidance Fusion Module (GFM), which are based on the reliable weights of the three modalities. Finally, we fuse these features to generate robust multi-modal features}
    \label{fig_third_case}
\end{figure*}

\subsection{Uncertainty in Deep Learning}
Uncertainty estimations~\cite{b, uncertainty1, uncertainty2, uncertainty3} have been introduced to deep learning models to identify areas where networks are error-prone. Lou \textit{et al.}~\cite{object} score uncertainties and propose a fusion module that exploits the Mixture-of-Expert architecture to encode multi-modal uncertainties in fusion pipelines to better percept autonomous driving. In the domain of object ReID, Zheng \textit{et al.}~\cite{Uncertainty_reid} reweight the contribution of losses based on uncertainty measured by output consistency; Yang \textit{et al.}~\cite{uncertainty_2} optimize the ReID model by selecting credible IDs to correct the noise in pseudo-labels. Furthermore, Zhao \textit{et al.}~\cite{uncertainty_3} model image and text features as Gaussian distributions and estimating their multi-granularity uncertainty via batch-level and identity-level variances which acted as a feature augmentation and provided richer image-text semantic relationship. The introduction of multi-modal is accompanied by the problem of multi-modal fusion. When researching multi-modal fusion methods, researchers should first focus on whether multi-modal fusion decisions are necessarily more effective than the previous single-modality or dual-modality approaches. Qiu \textit{et al.}~\cite{Does} indicate the importance of multi-modal learning in evaluating modality weights and thus measuring modality contributions when modality quality is unbalanced. Han \textit{et al.}~\cite{TMC} introduce a uncertainty-aware fusion framework based on Dirichlet distribution and Dempster-Shafer Theory to enhance both classification reliability and robustness. These methods collectively demonstrate that modeling uncertainty not only improves robustness under modality corruption but also enhances interpretability in multi-modal fusion. Compared with previous methods for calculating confidence in ReID tasks, our method of learning credible fusion incorporates the credibility of multi-modal when computing the weight of each modality. Thus, the reliable weights provide a foundation for us to classify multi-modal data with uncertain quality distributions, conduct targeted fusion while avoiding fusion conflicts, and fully leverage multi-modal data to improve ReID performance.

\section{METHODOLOGY}
To enable the vehicle ReID model to reliably and adaptively select processing strategies for different complex real-world scenarios, we ensure the processing after adaptive selection is targeted to these scenarios. As shown in Fig.~\ref{fig_third_case}, we design DCG is composed of three main components: Dynamic Confidence-based Disentangle Weighting (DCDW), Guidance Fusion Module (GFM) and Collaboration Fusion Module (CFM). There is  a sub-module (Inter-Mining) for achieving the interaction and mining of modal information. We will introduce these modules respectively in the following.

\subsection{Feature Extraction}
To effectively extract multi-modal features and reduce both modality differences and model parameters simultaneously, we employ a shared visual encoder from CLIP~\cite{CLIP} pre-trained on ViT-B/16 to process the RGB, NIR, and TIR features.
\begin{equation}
F_m = \Theta(I_m),
\end{equation}
where \( I_m \in \mathbb{R}^{H \times W \times C} \) denotes the input image from modality \( m \), and \( \Theta \) represents the shared vision encoder. The output token features \( F_m \in \mathbb{R}^{D \times (M+1)} \) are extracted from the final layer of the encoder, where \( D \) is the feature dimension and \( M \) is the number of patch tokens.

The output feature \( F_m \) consists of a class token and a set of patch tokens:
\begin{equation}
F_m = [F_m^{cls}, F_m^{pat}],
\end{equation}
where \( F_m^{cls} \) is the class token that captures global semantic information, and \( F_m^{pat} \) represents the local patch-level features.

\subsection{Dynamic Confidence-based Disentangling Weighting}
To differentiate fusion conflict data and assign a targeted fusion method to each sample more reliably, it is necessary to obtain reliable weights based on multi-modal interactions. In this case, we design a Dynamic Confidence-based Disentangling Weighting (DCDW) based on multi-modal confidence to comprehensively consider the reliable relationships among modalities. 

We calculate two distinct confidence metrics for each modality, \(Mono\)-\(Confidence\) and \(Holo\)-\(Confidence\), which accesses each modality's independent reliability and its collaborative reliability through multi-modal interactions, respectively. As Cao \textit{et al.}~\cite{predict} theoretically demonstrated, it better than the individual term to represent modality weight, we comprehensively consider two confidence measures, denoted as \(Co\)-\(Belief\), which reflects the confidence of every modality~\cite{Corbiere2019Addressing}. 

Firstly, we utilize \( F_m^{cls} \) , which has global features, pay more attention to local features, we first use \( F_m^{cls} \) as Q and \( F_m^{pat} \) as K and V. 
\begin{equation}
F_{I} = \Phi(Q_{cls}, K_{pat}),
\end{equation}
where $\Phi$ denotes the Multi-head Cross-attention~\cite{attn}.

Then, we calculate the confidence of the inter modality itself as \(Mono\)-\(Confidence\) scores for three modalities are computed as:



\begin{equation}
    M_{i,i\in{R,N,T}} = F_{I} \phi_{i},
\end{equation}
where $M_{i,i\in{R,N,T}}$ denotes $Mono\text{-}Confidence_{i,i\in{R,N,T}}$, $\phi_{i}$ are learnable parameters implemented with MLPs.

Secondly, we utilize \(Holo\)-\(Confidence\) to extend mono-confidence, rebalancing the contributions of the multi-modal. The \(Holo\)-\(Confidence\) for each modality is computed as:
\begin{equation}
 H_{i,i\in{R,N,T}} = \frac{\sum_{\substack{j \neq i}}^{N} M_{i,i\in{R,N,T}}}{\sum_{i=1}^{N} M_{i,i\in{R,N,T}}},
\end{equation}
where $H_{i,i\in{R,N,T}}$ denotes $Holo\text{-}Confidence_{i,i\in{R,N,T}}$, $j$ identifies the current modality, and \(N\) = 3.\\
The coherent belief $Co$-$belief$ incorporated two confidences is computed as:
\begin{equation}
    C_{i} = M_{i} + H_{i},
\end{equation}
where $C_{i}$ denotes $Co\text{-}belief_{i}$.

Finally, the dynamic multi-modal weights for each of the three modalities \(w_{R}\), \(w_{N}\) and \(w_{T}\) are computed as:
\begin{equation}
    w_{i,i\in{R,N,T}} = softmax( C_{i}).
\end{equation}

After obtaining multi-modal weights, it is possible to dynamically differentiate data with uncertain quality distributions for subsequent targeted interactive fusion without fusion conflict.
\subsection{Collaboration Fusion Module}
To enhance intra-class consistency when modal quality is relatively balanced, and mutually explore inter-modal features to amplify the gain of multi-modal interaction, we introduce Collaboration Fusion Module (CFM). 
In the experiments of ReID with missing modalities~\ref{table:missing}, we found that in some cases, the performance metrics were higher when certain modalities were missing than when all modalities were present. We realized that in situations with varying challenges and uneven quality, forcibly fusing poor-quality modalities is not as effective as discarding them. Therefore, we design a sample-level approach to discard a certain poor-quality modality for a specific sample.

In certain case, one modality may exhibit severely degraded quality such as RGB under low-light or night scenarios, making tri-modal fusion may introduce interference rather than complementary information. 

So while conducting collaborative fusion, we introduce the modality retention mechanism:
\begin{equation}
\tau = \sigma\left( \zeta\left( W_{\text{avg}} \right) \right),
\end{equation}
where  $\zeta$ denotes MLP, \(W_{avg}= (w_{R}+w_{N}+w_{T})/3\), \(w_{R}\), \(w_{N}\) and \(w_{T}\) are obtained from the DCDW, \(\sigma\) is the Sigmoid function (constraining \(\tau \in [0, 1]\)) and MLP models the statistical characteristics of modal weights. We initialize \(w_{T}\) to 0.25 and design it as a learnable parameter. Under the setting where the sum of the three-modal weights is 1, the model will not encounter a situation where all three-modal samples are discarded simultaneously.

A modality m is discarded if its confidence weight falls below \(\tau\):
\begin{equation}
\mathcal
{D}(m) =
\begin{cases}
1, &\text{if}~w_m < \tau \quad (\text{discard}),
\\0, &\text{if}~w_m \geq \tau \quad (\text{retain}),
\end{cases}
\end{equation}
\begin{equation}
    M_{retain} = {m|D(m) = 0},
\end{equation}
where $M_{retain}$ denotes retained modalities.

\noindent\textbf{Inter-Mining.} Based on discarding low-quality samples, we design a bidirectional mining module named Inter-Mining to explore the consensus information among modalities to interactively mine inter-class modality features.

Specifically, each pair of the retained modalities acts as a query to mine feature information from the other modality. For each modality, we use $F_{m}$ as query $Q_{m}$ to mine the features of $K_{n}$, where m $\neq$ n and m, n $\in$ \{R, N ,T\}:
\begin{equation}
    F_{m2n} = \Phi(Q_{m}, K_{n}),
\end{equation}
where $\Phi$ denotes the Multi-head Cross-attention ~\cite{attn}.

To enhance feature representation~\cite{cvpr21}, the features are split into two branches for separate processing, which is equivalent to enhancing features from different perspectives \(F_{1}^{m2n}\) and \(F_{2}^{m2n}\):
\begin{equation}
    F_{1}^{m2n}, F_{2}^{m2n} = Split(DWConv_{3}(F_{m2n}),dim=1).
\end{equation}
Then, the Tanh function constrains the value of features in the range of [-1, 1], enhancing the contrast of the features and improving the robustness of the features against changes in illumination~\cite{HVI}.
\begin{align}
    F_{1}^{'mn} &= T(D(F_{1}^{mn})) + F_{1}^{m2n},\\
    F_{2}^{'mn} &= T(D(F_{2}^{mn})) + F_{2}^{m2n},
\end{align}
where $T$ denotes Tanh function, and $D$ denotes DWconv.

In order to encourage interaction between the two enhanced branches and highlight regions that are discriminative in both branches, we perform element-wise multiplication on features from two perspectives.
\begin{equation}
    F_{m2n}^{'} =  F_{1}^{'mn}\odot F_{2}^{'mn} + F_{n},
\end{equation}
where \(\odot\) means element-wise multiplication.

Finally, we obtain the interactive mined features \(F_{m2n}^{'}\). We fuse these interaction results to enhance the multi-modal feature representation:
\begin{equation}
    F_{all} = \zeta(Concat(F_{m2n}^{'}, F_{n2m}^{'})),
\end{equation}
where $\zeta$ denotes MLPs.
Based on confidence-based dropout and inter-mining of useful features, we enhance intra-class consistency thereby fully realizing the complementary multi-modal advantages. 
\subsection{Guidance Fusion Module}
When modal weight distributions are imbalanced, which we set a learnable parameter $\beta$: $w_{i,i\in{rgb, nir, tir}}$ $>$ $\beta$ (inicialized as 0.35). In this case, we design a Guidance Fusion Module (GFM) grounded in Bayesian uncertainty theory~\cite{b} to amplify dominant modality advantages and enable cross-modal complementarity. 

Firstly, the dominant modality is selected by integrating epistemic uncertainty with modal weights:
\begin{equation}
    \mathcal{M}_{\text{dom}} = \arg\max_{m \in \{R,N,T\}} \left( \mathcal{U}_e(m) \cdot W(m) \right)
\end{equation}
where $\mathcal{U}_e(m)$ denotes the epistemic uncertainty of modality $m$ (estimated via Monte Carlo dropout \cite{b}), and $W(m)$ represents modal weights, adhering to Bayesian decision principles.
Secondly, we compute spatial-aware discrepancy vectors:
\begin{equation}
    d_{xy}^{\text{conv}} = \text{SSM}\left( \zeta(d_{xy}) \right),
\end{equation}
where $\zeta$ denotes Conv, $d_{xy}$ = $F_{\text{dom}}$ - $F_j$ (with $j$ as auxiliary modalities), and $\text{SSM}$ denotes spatial attention to emphasize discriminative patterns in unique features. \\
These discrepancies are then amplified via learnable factors, we use $F_{\text{dom}}$ as reference:
\begin{equation}
    F_{\text{dom}}^{\text{En}} = F_{\text{dom}} + \alpha_{ij} \cdot d_{ij} + \alpha_{iz} \cdot d_{iz},
\end{equation}
where $\alpha_{ij}, \alpha_{iz} \in \mathbb{R}^D$ are learnable amplification factors. By maximizing $d_{xy}$, the model implicitly separates shared and unique information, aligning with the invariance principle in representation learning \cite{b2}.

Thirdly, cross-modal guidance propagates dominant modality information to auxiliaries. Enhanced auxiliary features are obtained via:
\begin{equation}
    \text{Attn}_{j \leftarrow \text{dom}} = \text{Softmax}\left( \frac{\text{Cos}(F_{\text{dom}}^{\text{En}}, F_j)}{\sqrt{d}} \right),
\end{equation}
\begin{equation}
    F_j^{\text{En}} = F_j + \text{Attn}_{j \leftarrow \text{dom}} \odot d_{xy},
\end{equation}
where Cos(,) denotes Cosine Similarity~\cite{cos}, $\odot$ denotes element-wise multiplication, ensuring uncertainty-aware guidance.

Finally, multi-modal fusion aggregates interactive features:
\begin{equation} 
    \text{inter}_{j2i} = \zeta\left( Concat(F_j^{\text{En}}, F_i^{\text{En}}) \right), \quad 
\end{equation}
\begin{equation} 
    F_{\text{All}} = Concat\left( \text{inter}_{j2i}, \text{inter}_{k2i}, F_i^{\text{En}} \right), 
\end{equation} 
where $\zeta$ denotes Conv, $i$ = $\arg\max W(m),\ \{j,k\} = \{R,N,T\} \setminus \{i\}$.

This framework generalizes to any dominant modality identified by weight distribution, ensuring consistent amplification of dominant advantages and interactive mining of non-dominant information across varying quality distributions. Through modality interactive mining based on amplifying the dominance of high-confidence modalities, it achieves comprehensive utilization of modal information, narrowing modality heterogeneity while enhancing intra-class consistency.

\subsection{Objective Function}
During the training phase, the backbone features and the fusion features of our DCG are jointly supervised by identity loss~\cite{idLOSS} and triplet loss~\cite{tripletloss}.
\begin{equation}
    L_{ReID}(X) = L_{id}(X) + L_{tri}(X),
\end{equation}
where $X$ represents input features for supervision. Finally, the overall loss for our framework is defined as:
\begin{equation}
    L_{total} = L_{ReID}^{DCG}(f) + L_{ReID}^{CLIP}([f_{R}, f_{N}, f_{T}]).
\end{equation}

\section{EXPERIMENT}
In this section, we conduct detailed experiments on the
proposed framework. First, we introduce the datasets, evaluation protocols, and implementation details (Sec.IV.A-B). Second, we conduct comparative experiments with the latest methods on vehicle datasets and we conduct comparative experiments with missing modalities (Sec.IV.C). Then, we perform ablation experiments (Sec.IV.D) including effects of key components and discussions of our modules. Finally we further analyze the components of the framework through visual analysis of the proposed method (Sec.IV.E). 
\begin{table*}[ht!]
\renewcommand\arraystretch{1.15} 
\centering
\caption{Performance comparison on multi-modal vehicle ReID datasets. The best results are in \textbf{bold} and the second in \underline{underlined}. The symbol $\dagger$ denotes CLIP-based methods, $*$ indicates ViT-based methods, and others are CNN-based methods.}
\footnotesize 
\begin{tabularx}{\textwidth}{c@{\hspace{0.2em}}c@{\hspace{0.2em}}c@{\hspace{0.2em}}>{\centering\arraybackslash}X|c|cc|cccc|cccc}
\hline
\multicolumn{4}{c|}{\multirow{2}{*}{\textbf{\small Methods}}} & \multirow{2}{*}{\textbf{\small Venue}} & \multicolumn{2}{c|}{\textbf{RGBNT100}} & \multicolumn{4}{c|}{\textbf{WMVeID863}} & \multicolumn{4}{c}{\textbf{MSVR310}} \\ \cline{6-15} 
\multicolumn{4}{c|}{} & & \textbf{mAP} & \textbf{R-1} & \textbf{mAP} & \textbf{R-1} & \textbf{R-5} & \textbf{R-10} & \textbf{mAP} & \textbf{R-1} & \textbf{R-5} & \textbf{R-10} \\ \hline
\multirow{4}{*}{{\makecell{\textbf{{\rotatebox{90}{\textbf{Single}}}}}}} 
& & & BoT~\cite{BoT}         & CVPRW'19 & 78.0 & 95.1 & 51.1 & 55.7 & 69.8 & 74.7 & 23.5 & 38.4 & 56.8 & 64.8 \\
& & & OSNet~\cite{osnet}      & ICCV'19  & 75.0 & 95.6 & 42.9 & 46.8 & 61.9 & 69.4 & 28.7 & 44.8 & 66.2 & 73.1 \\
& & & AGW~\cite{AGW}          & TPAMI'21 & 73.1 & 92.7 & 30.3 & 35.3 & 43.3 & 46.5 & 28.9 & 46.9 & 64.3 & 70.7 \\
& & & TransReID$^*$~\cite{transreid} & ICCV'21  & 75.6 & 92.9 & 67.0 & 74.7 & 79.5 & 82.4 & 26.9 & 43.5 & 62.4 & 70.7 \\ \hline
\multirow{16}{*}{{\makecell{\textbf{\rotatebox{90}{\textbf{Multi-modal}}}}}} 
& & & HAMNET~\cite{HAMNET}    & CVPR'18  & 74.5 & 93.3 & 45.6 & 48.5 & 63.1 & 68.8 & 27.1 & 42.3 & 61.6 & 69.5 \\
& & & PFNet~\cite{PFNET}      & AAAI'21  & 68.1 & 94.1 & 50.1 & 55.9 & 68.7 & 75.1 & 23.5 & 37.4 & 57.0 & 67.3 \\
& & & IEEE~\cite{IEEE}        & AAAI'22  & 61.3 & 87.8 & 45.9 & 48.6 & 64.3 & 67.9 & 21.0 & 41.0 & 57.7 & 65.0 \\
& & & CCNet~\cite{CCNET310}   & INFFU'23 & 77.2 & 96.3 & 50.3 & 52.7 & 69.6 & 75.1 & 36.4 & 55.2 & 72.4 & 79.7 \\
& & & TOP-ReID$^*$~\cite{top}  & AAAI'24  & 81.2 & 96.4 & 67.7 & 75.3 & 80.8 & 83.5 & 35.9 & 44.6 & -    & -    \\
& & & EDITOR$^*$~\cite{Editor} & CVPR'24  & 82.1 & 96.4 & 65.6 & 73.8 & 80.0 & 82.3 & 39.0 & 49.3 & -    & -    \\
& & & HTT$^*$~\cite{HTT} & AAAI'24  & 75.7 & 92.6 & 66.2 & 73.2 & 79.9 & 82.3 & 34.5 & 43.2 & -    & -    \\
& & & FACENet$^*$~\cite{Facenet} & INFFU'25 & 81.5 & 96.9 & 69.8 & 77.0 & 81.0 & 84.2 & 36.2 & 54.1 & -    & -    \\ \cline{4-15} 
& & & DeMo$^\dagger~$~\cite{Demo} & AAAI'25  & 86.2 & \underline{97.6} & 68.8 & 77.2 & 81.5 & 83.8 & 49.2 & 59.8 & -    & -    \\
& & & Mambapro$^\dagger~$~\cite{mambapro} & AAAI'25 & 83.9 & 94.7 & 69.5 & 76.9 & 80.6 & 83.8 & 47.0 & 64.0 & -    & -    \\
& & & IDEA$^\dagger~$~\cite{IDEA:} & CVPR'25  & \underline{87.2} & 96.5 & - & - & - & - & 47.0 & 62.4 & -    & -    \\
& & & PromptMA$^\dagger~$~\cite{promptMA} & TIP'25 & 85.3 & 97.4 & - & - & - & - & 55.2 & 64.5 & -    & -    \\
& & & CoEN$^\dagger~$~\cite{coen} & arxiv'25 & 86.4 & 96.6 & \underline{70.9} & \underline{77.9} & \underline{82.7} & \underline{85.6} & 52.2 & 69.5 & - & - \\
& & & ICPL$^\dagger~$~\cite{ICPL-ReID} & TMM'25 & 87.0 & \textbf{98.6} & 67.2 & 74.0 & 81.2 & 85.6 & \underline{56.9} & \textbf{77.7} & \underline{87.6} & \underline{91.5} \\
& & & \textbf{DCG}$^\dagger~$ & \textbf{Ours} & \textbf{87.6} & \underline{97.6} & \textbf{71.4} & \textbf{80.1} & \textbf{85.4} & \textbf{88.6} & \textbf{62.9} & \underline{77.5} & \textbf{91.7} & \textbf{94.2} \\ \hline
\end{tabularx}
\label{tab:performance on vehicle datasets}
\end{table*}
\subsection{Datasets and Evaluation Protocols}
We conduct experiments on three publicly available multi-spectral vehicle ReID datasets MSVR310~\cite{CCNET310}, RGBNT100~\cite{100}, and WMVeID863~\cite{Facenet} which include three vehicle scenarios with different challenges.

\noindent\textbf{WMVeID863~\cite{Facenet}}
contains 4709 image triplets of 863 IDs of 8 camera
views, the number of image triplets of each vehicle varies from 1 to 39. The datasets are randomly selected 575 IDs with 3314 image triplets for training and 288 IDs with 1309 image triplets for testing. The gallery samples consist of all 288 IDs with 1309 image triplets. The query samples are randomly selected from the 288 gallery samples, consisting of 205 IDs with 928 image triplets.

\noindent\textbf{MSVR310~\cite{CCNET310}}
contains 6,261 high-quality vehicle images, divided into 310 different vehicles, with 2,087 samples, each consisting of 3 spectra modalities. The training set includes 155 vehicles and a total of 1,032 samples. The gallery set contains 1,055 samples of the remaining 155 vehicles, while the query set consists of 52 randomly selected vehicles and 591 samples from the gallery set. These samples are captured at long time spans, covering 8 viewpoints around the vehicle and various challenges such as illumination change, shadow,
reflection, and color distortion.

\noindent\textbf{RGBNT100~\cite{100}}
selected 100 vehicles and added 17,250 additional thermal-infrared images to form a three-spectra dataset. This dataset includes 8,675 image triples
from 50 vehicles for the training set and 8,575 triples from the other 50 vehicles for the test gallery set. From the test gallery, 1,715 samples are selected to form the query set.

\noindent\textbf{Evaluation Protocols}
To ensure fair experimental evaluation, we follow previous research methods. We use mean Average Precision(mAP) which measures the average precision of all queries, represented by the area under the precision-recall curve, providing a comprehensive assessment of recall and precision performance and Cumulative Matching Characteristics (CMC) at Rank-K (K = 1, 5, 10) to assess performance, commonly used in ReID tasks, to evaluate our approach.

\subsection{Implementation Details}
Our model is implemented with the Pytorch toolbox. We conduct experiments on one NVIDIA RTX 3090 GPU. We resize the images of each spectrum to $128 \times 256$ in vehicle datasets to maintain the aspect ratio of vehicle and use random horizontal flipping, padding with 10 pixels, random cropping, and random erasing~\cite{random} as feature enhancement strategies. We present trainable parameters and FLOPs for complexity analysis. We select ViT-B/16 as our visual backbone.
We optimize the model via Adam: initial learning rates \(2.5\times{10^{-4}}\) for WMVeID863~\cite{Facenet} , \(1\times{10^{-5}}\) for RGBNT100~\cite{100} and \(3.5\times{10^{-3}}\) for MSVR310~\cite{CCNET310}. 
\subsection{Comparison with State-of-the-art Methods}
\noindent\textbf{Comparison on WMVeID863~\cite{Facenet}.}
As shown in the Table.~\ref{tab:performance on vehicle datasets}, we compare our method with the current state-of-the-art single-spectral and multi-spectral approaches. Notably, on the WMVeID863~\cite{Facenet} dataset, which introduces severe illumination interference caused by strong lens flares, our method DCG achieves the best performance across all evaluation metrics. To demonstrate the superiority of our method through comparison, we compare it with excellent methods specifically designed for vehicle challenges. Compared with FACENet, which enhances other spectras by leveraging the flare immune characteristics of the TIR spectrum, DCG outperforms it by~\textbf{+1.6\%} in mAP and~\textbf{+3.1\%} in Rank-1. In FACENet, KL divergence is used to constrain the semantic consistency of modalities. However, in cases where the quality of multi-modal is uneven, it is prone to reducing the discriminative power of the fused features when the RGB is contaminated by glare. Compared with CoEN, a framework that can adaptively select and enhance the dominant modality to address the issue of modal imbalance, our DCG disentanging fusion method outperforms it in terms of both mAP and Rank-1. Although CoEN considers using the dominant mode to assist in optimizing the secondary mode to enhance the model's recognition ability under the condition of uneven quality distribution, it has not yet addressed the situation where the three modes have balanced mass. These demonstrate the applicability of our DCG to uncertain modality quality distributions caused by complex vehicle scenarios.
\begin{table*}[htbp]
\vspace*{-\baselineskip} 
\centering
\caption{Performance of missing-modality settings on MSVR310~\cite{CCNET310}, WMVeID863~\cite{Facenet} and RGBNT100~\cite{100}. ``M(X)'' means missing the X image modality. The best results are in \textbf{bold} and the second in \underline{underlined}.}
\setlength{\abovecaptionskip}{3pt}
\resizebox{\textwidth}{!}{
\begin{tabular}{ccccccccccccccc} 
\toprule
\multirow{2}{*}{\textbf{Methods}} & \multicolumn{2}{c}{\textbf{M (RGB)}} & \multicolumn{2}{c}{\textbf{M (NIR)}} & \multicolumn{2}{c}{\textbf{M (TIR)}} & \multicolumn{2}{c}{\textbf{M (RGB+NIR)}} & \multicolumn{2}{c}{\textbf{M (RGB+TIR)}} & \multicolumn{2}{c}{\textbf{M (NIR+TIR)}} & \multicolumn{2}{c}{\textbf{Average}} \\
\cmidrule(lr){2-3} \cmidrule(lr){4-5} \cmidrule(lr){6-7} \cmidrule(lr){8-9} \cmidrule(lr){10-11} \cmidrule(lr){12-13} \cmidrule(lr){14-15}
& \textbf{mAP} & \textbf{R-1} & \textbf{mAP} & \textbf{R-1} & \textbf{mAP} & \textbf{R-1} & \textbf{mAP} & \textbf{R-1} & \textbf{mAP} & \textbf{R-1} & \textbf{mAP} & \textbf{R-1} & \textbf{mAP} & \textbf{R-1} \\
\midrule
\multicolumn{15}{c}{\textbf{RGBNT100}} \\
\midrule
CCNet~\cite{CCNET310} & 66.8 & 90.2 & 73.2 & 92.2 & 60.0 & 82.9 & 44.4 & 75.0 & 42.4 & 63.8 & 49.5 & 69.8 & 56.0 & 79.0 \\
TOP-ReID~\cite{top} & 70.6 & 90.6 & 77.9 & 94.5 & 64.0 & 81.5 & 42.5 & 69.3 & 45.9 & 65.4 & 55.4 & 77.8 & 59.4 & 79.9 \\
DeMo~\cite{Demo} & \textbf{81.0} & 94.1 & \underline{84.1} & 96.5 & \textbf{71.1} & \underline{87.6} & \underline{50.2} & 73.7 & \textbf{59.6} & \textbf{78.1} & \underline{66.3} & \underline{82.8} & \underline{68.7} & \underline{85.5} \\
PromptMA~\cite{promptMA} & 80.3 & \textbf{95.8} & 83.5 & \textbf{96.8} & 69.3 & 85.8 & 48.0 & \underline{74.3} & 56.2 & \underline{75.2} & 64.2 & 82.0 & 66.9 & 85.0 \\
\textbf{DCG (Ours)} & \textbf{81.7} & \underline{95.3} & \textbf{87.1} & \underline{96.5} & \underline{70.8} & \textbf{87.8} & \textbf{51.1} & \textbf{76.2} & \underline{58.4} & 74.8 & \textbf{69.1} & \textbf{85.4} & \textbf{69.7} & \textbf{86.0} \\
\midrule
\multicolumn{15}{c}{\textbf{MSVR310}} \\
\midrule
CCNet~\cite{CCNET310} & 25.5 & 43.7 & 27.2 & 42.3 & 26.4 & 41.6 & 10.8 & 30.1 & 20.6 & 34.2 & 24.1 & 37.2 & 22.4 & 38.2 \\
TOP-ReID~\cite{top} & 23.5 & 41.1 & 28.6 & 41.8 & 31.1 & 45.9 & 10.8 & 23.5 & 22.3 & 40.6 & 26.5 & 36.5 & 23.8 & 38.2 \\
DeMo~\cite{Demo} & \underline{36.9} & \underline{55.3} & \underline{43.1} & \underline{56.5} & \underline{46.1} & \underline{60.9} & \underline{10.5} & \underline{24.2} & \underline{34.1} & \underline{53.5} & \underline{40.8} & \underline{53.6} & \underline{35.5} & \underline{50.7} \\
\textbf{DCG (Ours)} & \textbf{54.2} & \textbf{71.0} & \textbf{59.3} & \textbf{69.9} & \textbf{57.2} & \textbf{67.8} & \textbf{28.1} & \textbf{39.9} & \textbf{45.0} & \textbf{60.1} & \textbf{53.4} & \textbf{63.4} & \textbf{49.5} & \textbf{62.0} \\
\midrule
\multicolumn{15}{c}{\textbf{WMVeID863}} \\
\midrule
CCNet~\cite{CCNET310} & 58.7 & 64.8 & 59.8 & 64.9 & 53.5 & 58.7 & 51.9 & 56.8 & 43.0 & 44.1 & 46.7 & 48.6 & 51.9 & 56.8 \\
TOP-ReID~\cite{top} & 63.2 & 69.9 & \underline{66.8} & \underline{73.7} & 55.3 & \underline{59.8} & \underline{60.3} & \underline{65.7} & 46.8 & \underline{49.5} & \underline{53.0} & \underline{56.4} & 57.6 & 62.5 \\
DeMo~\cite{Demo} & \underline{66.5} & \underline{73.3} & 65.9 & 73.3 & \underline{55.5} & \underline{59.8} & 58.4 & 63.9 & \underline{49.7} & 52.1 & 50.4 & 52.8 & \underline{57.7} & \underline{62.5} \\
\textbf{DCG (Ours)} & \textbf{70.6} & \textbf{78.1} & \textbf{73.1} & \textbf{79.9} & \textbf{61.4} & \textbf{68.7} & \textbf{64.3} & \textbf{71.0} & \textbf{57.0} & \textbf{63.0} & \textbf{58.7} & \textbf{63.5} & \textbf{64.1} & \textbf{70.7} \\
\bottomrule
\end{tabular}
}
\label{table:missing}
\end{table*}

\noindent\textbf{Comparison on RGBNT100~\cite{100} and MSVR310~\cite{CCNET310}.}
On the large-scale RGBNT100~\cite{100} dataset, IDEA constructs a text-enhanced multi-modal object ReID benchmark, leveraging modal prefixes and inverse networks to integrate multi-modal features, further refreshing the performance to 87.2\% mAP and 96.5\% Rank-1. However, DCG continues to set new benchmarks by achieving 87.6\% mAP and 97.6\% Rank-1 on the RGBNT100~\cite{100} dataset by decoupling fusion and two designed fusion strategies.
When faced with the challenges of viewpoint changes and long time spans in the MSVR310~\cite{CCNET310} dataset, common methods often experience performance degradation. In contrast, our DCG framework, designed specifically for complex and dynamic scenarios, outperforms DeMo and PromptMA on MSVR310~\cite{CCNET310} by \textbf{+13.7\%/+7.7\%} in mAP and \textbf{+17.7\%/+13\%} in Rank-1. This fully demonstrates the exceptional effectiveness of our fusion framework in handling significant variations in viewpoint and temporal span.

In summary, these results validate the superiority of the DCG method in the face of complex vehicle scenarios and large-scale vehicle datasets, proving that our method can address the main challenges of multi-modal vehicle ReID.

\noindent\textbf{Multi-modal Vehicle ReID with Missing Modalities.}
To assess DCG’s robustness in missing-modality scenarios, as shown in Table.~\ref{table:missing}, we conduct experiments on WMVeID863~\cite{Facenet}, MSVR310~\cite{CCNET310} and RGBNT100~\cite{100}. By comparing with advanced methods, our performance remains state-of-the-art even when there are missing modalities. Despite lacking specific designs like the reconstruction modules in TOP-ReID~\cite{top} and PromptMA~\cite{promptMA}, our DCG achieves competitive performance through dynamic multi-modal fusion. In terms of average performance across the three vehicle datasets with different challenges, our method achieves the best results in both mAP and Rank-1. Specifically, on MSVR310, its mAP/Rank-1 are \textbf{+14.0\%/+11.3\%} higher than those of DeMo~\cite{Demo} respectively. On the WMVeID863 dataset with vehicle flare pollution, the average mAP/Rank-1 are improved \textbf{+6.4\%/+8.2\%} compared with DeMo. On the large-scale dataset RGBNT100~\cite{100}, the average performance of our method in terms of mAP/Rank-1 is \textbf{+2.8\%/+0.5\%} higher than that of DeMo. Combined with the experimental data of multiple missing scenarios, it shows that our method can handle various modality missing problems well.

These results validate the superiority of the DCG approach in complex scenarios multi-spectral vehicle datasets, demonstrating its effectiveness over existing solutions.

\subsection{Ablation Study}
To thoroughly validate the effectiveness of each component of our proposed modules, we conduct ablation studies on the WMVeID863~\cite{Facenet} dataset on the Clip-based~\cite{CLIP} backbone to validate the proposed components.

\noindent\textbf{Effects of Key Components.}
Table.~\ref{tab:ablation_components} shows the performance comparison with different modules. Model A is the baseline model. Model B and Model C introduce two fusion methods GFM and CFM respectively, increasing the mAP/Rank-1 \textbf{+3.9\%/+4.8\%} and \textbf{+3.7\%/+4.6\%}, which indicates the effectiveness of our two fusion strategies. The model B lies in using learnable factors to amplify the value of the dominant modality and guiding auxiliary modalities to focus on effective information. This design perfectly adapts to the characteristic of multi-modal data where a certain modality gains an advantage due to specific scene conditions, thereby verifying the effectiveness of the module in fusion. When the quality of multi-modal vehicle data is balanced, the information carried by each modality exhibits strong complementarity. The model C introduces CFM leverages confidence-based dropout to retain the core information of each modality while discarding redundant contour features. Subsequently, it employs bidirectional attention to enforce information alignment across modalities. The experimental results of model C demonstrate the effectiveness of this design. And then, to fully validate the effectiveness of our DCDW, we replace our DCDW selected weights by randomly feeding modal samples into the two fusion strategies formed Model D. Compared to our confidence-based differentiation method DCDW, the mAP/Rank-1 perform \textbf{-1.6\%/-1.8\%} and \textbf{-1.1\%/-2.9\%}, respectively. Model E combines the DCDW and two fusion strategies, achieving the best results with mAP \textbf{71.1\%} and Rank-1 \textbf{80.1\%}. These results clearly demonstrate the effectiveness and efficiency of our proposed modules. 
\begin{table}[htbp]
\setlength{\abovecaptionskip}{3pt}  
\setlength{\belowcaptionskip}{6pt}  
\centering
\caption{Performance comparison with different modules.}
\normalsize  
\setlength{\tabcolsep}{2.8pt}  
\begin{tabular*}{\columnwidth}{@{\extracolsep{\fill}} c c c c | c c c c @{}}
\hline
\multicolumn{4}{c|}{\textbf{Component}} & \multicolumn{4}{c}{\textbf{WMVeID863}} \\ \hline
\multirow{2}{*}{\textbf{Index}} & \multirow{2}{*}{\textbf{DCDW}} & \multicolumn{2}{c|}{\textbf{Fusion}} & \multirow{2}{*}{\textbf{mAP}} & \multirow{2}{*}{\textbf{R-1}} & \multirow{2}{*}{\textbf{R-5}} & \multirow{2}{*}{\textbf{R-10}} \\ \cline{3-4}
 &  & \textbf{GFM} & \textbf{CFM} &  &  &  &  \\ \hline
A & × & × & × & 66.7 & 74.2 & 80.6 & 82.3 \\
B & \checkmark & \checkmark & × & 70.6 & 79.0 & 84.7 & 86.7 \\
C & \checkmark & × & \checkmark & 70.4 & 78.8 & 83.5 & 85.9 \\
D & × & \checkmark & \checkmark & 69.8 & 78.3 & 84.0 & 86.5 \\
\textbf{E} & \textbf{\checkmark} & \textbf{\checkmark} & \textbf{\checkmark} & \textbf{71.1} & \textbf{80.1} & \textbf{84.9} & \textbf{86.7} \\ \hline
\end{tabular*}
\label{tab:ablation_components}
\end{table}
\begin{table}[htbp]
\centering
\caption{Comparative discussion on dynamic sample partitioning methods of WMVeID863~\cite{Facenet} dataset. Among them, w/o (free) indicates that samples are randomly partitioned; w/o (TMC) indicates that we partition samples using the evidence theory~\cite{TMC} method; w/o (all) indicates that we treat samples uniformly and feed them into both fusion methods simultaneously.}
\normalsize  
\setlength{\tabcolsep}{3.5pt}  
\begin{tabular*}{\columnwidth}{@{\extracolsep{\fill}} c c c c c c @{}}  
\hline
\multirow{2}{*}{\textbf{Index}} & \multirow{2}{*}{\textbf{Method}} & \multicolumn{4}{c}{\textbf{WMVeID863}} \\ \cline{3-6} 
& & \textbf{mAP} & \textbf{R-1} & \textbf{R-5} & \textbf{R-10} \\ \hline
A & w/o (free)       & 69.8 & 78.3 & 84.0 & 86.5 \\
B & w/o (TMC)        & 69.2 & 75.3 & 84.2 & 86.9 \\
C & w/o (all)        & 70.3 & 77.2 & 82.0 & 85.1 \\
\textbf{D} & \textbf{w (ours)}    & \textbf{71.4} & \textbf{80.1} & \textbf{85.4} & \textbf{88.6} \\ \hline
\end{tabular*}
\label{tab:DCDW}
\end{table}
\begin{figure}[htbp]  
\hfill  
\includegraphics[width=1.0\linewidth]{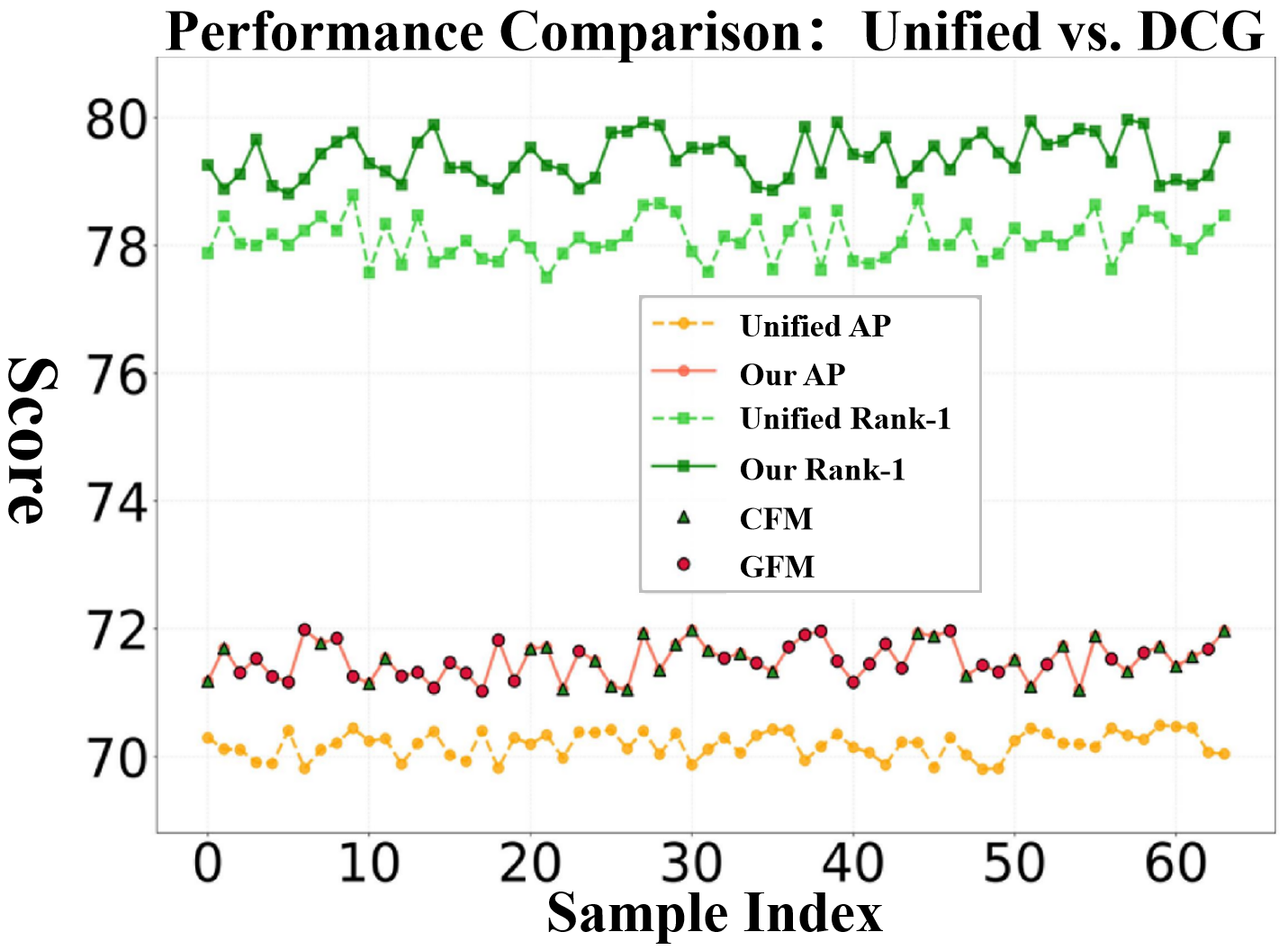}  
\caption{Comparison of WMVeID863 performance on samples between our decoupling fusion method and unified fusion method (feeding all samples into two fusion strategies).}
\label{fig:comparison}
\end{figure}

\noindent\textbf{Discussion on DCDW.}
As shown in Table.~\ref{tab:DCDW}, we evaluate the effectiveness of DCDW. Specifically, when we randomly select fusion strategies for samples (A), the mAP/Rank-1 scores decrease by \textbf{-1.6\%/-1.8\%} respectively. To explore other methods for evaluating sample quality in the application of ReID, as in Table.~\ref{tab:DCDW} model B, we compare the TMC~\cite{TMC} which is based on evidence theory with our DCG. Our DCDW outperforms the evidence theory-based approach on WMVeID863 by leveraging both the modality itself and the interaction between modalities. TMC treats each modality as an independent evidence source to improve the accuracy of category decision, but DCG evaluates modal quality by integrating both the modalities themselves and their interactions to make fusion decisions. Our goal of using confidence assessment to evaluate the quality of modalities is to distinguish the quality distributions of multi-modal data. Through our decision-making approach that considers both the modalities themselves and their interactions, DCDW achieves a better way of classifying multi-modal fusion decisions. We feed all samples into two fusion strategies (C) to verify the necessity of fusion decoupling. When samples are subjected to two fusion measures simultaneously, compared with our fusion decoupling framework, the mAP/Rank-1 scores decrease by \textbf{-1.1\%/-2.9\%}, respectively. By comparing several different methods of assigning multi-modal data to fusion measures with our confidence-based reliable assignment, we can verify the reliability of our approach. The confidence-based dynamic allocation mechanism we designed for multi-modal data, which directs data into targeted fusion methods, exhibits higher reliability in handling the quality distribution characteristics of multi-modal data. 
\begin{figure*}[htbp]  
\centering  
\includegraphics[width=1.0\textwidth]{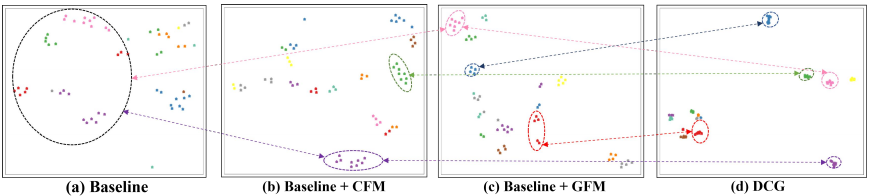}  
\caption{Feature distributions with t-SNE~\cite{tsne} on WMVeID863. Different colors represent different identities. (a) Baseline, (b) Baseline + CFM, (c) Baseline + GFM, (d) DCG. Better
view with colors and zooming in. }  
\label{fig:tsne}  
\end{figure*}

\noindent\textbf{Discussion on Fusion Decoupling.}
As shown in  Fig.~\ref{fig:comparison}, we use line charts and graphic markers to more clearly demonstrate the effectiveness of fusion and decoupling, while comparing our method with the one that does not adopt fusion and decoupling. For each multi-modal sample, the AP and Rank-1 values obtained when processing through both fusion methods simultaneously are lower than those achieved by our fusion decoupling approach. This directly demonstrates the effectiveness of our method. These experiments fully demonstrate our DCDW can avoid fusion conflicts and the necessity of fusion decoupling. By comparing several different methods of assigning multi-modal data to fusion measures with our confidence-based reliable assignment, we can verify the reliability of our approach. The confidence-based dynamic allocation mechanism we designed for multi-modal data, which directs data into targeted fusion methods, exhibits higher reliability in handling the quality distribution characteristics of multi-modal data. By comparing several different methods of assigning multi-modal data to fusion measures with our confidence-based reliable assignment, we can verify the reliability of our approach. The confidence-based dynamic allocation mechanism we designed for multi-modal data, which directs data into targeted fusion methods, exhibits higher reliability in handling the quality distribution characteristics of multi-modal data.
\begin{table}[htbp]
\setlength{\abovecaptionskip}{3pt}  
\setlength{\belowcaptionskip}{6pt}  
\centering
\caption{Comparison of removing dropout (Drop) and differential amplification (Amplify) in CFM and GFM with the full model.} 
\resizebox{1.0\linewidth}{!}{  
\begin{tabular}{c|cccc}  
\hline
\multirow{2}{*}{\textbf{Methods}} & \multicolumn{4}{c}{\textbf{WMVeID863}}                        \\ \cline{2-5} 
                                   & \textbf{mAP}  & \textbf{R-1}  & \textbf{R-5}  & \textbf{R-10} \\ \hline
w/o CFM (Drop)                      & 70.8          & 78.3          & 82.0          & 85.1          \\
w/o GFM (Amplify)                   & 70.3          & 77.0          & 79.8          & 83.2          \\
\textbf{DCG (ours)}                & \textbf{71.4} & \textbf{80.1} & \textbf{85.4} & \textbf{88.6} \\ \hline
\end{tabular}}
\label{tab:DETAIL}
\end{table}

\noindent\textbf{Discussion of CFM and GFM.} As shown in the Table.~\ref{tab:DETAIL} for CFM, we remove its dropout design, meaning that instead of discarding modalities with low weights, we treat all modalities equally and perform pairwise feature mining. When the quality of multi-modal data is balanced, we have incorporated a design to discard the low-quality modality in scenarios where one modality exhibits poor quality while the other two maintain comparable quality. Based on a learnable discard factor, we can adaptively discard the low-quality modality, thereby avoiding the negative impact of this modality on the fusion of multi-modal data. Compared with DCG, mAP/Rank-1 perform \textbf{-0.6\%/-1.8\%} respectively, which verifies the rationality of our design in CFM to discard the modality with low weights. As shown in the Table.~\ref{tab:DETAIL} for GFM, we remove the factor \(\alpha\) in the GFM that amplifies the dominant modality, leaving the dominant modality with only a guiding function. When the quality of multi-modal data is unbalanced, there will be a dominant modality. A real-world scenario corresponding to this situation is that, at night, RGB imaging quality is poor while NIR imaging quality is excellent. The adaptive amplification mechanism of GFM can effectively amplify the advantages of the dominant modality, thereby enhancing the fusion performance of multi-modal data. As a result, mAP/Rank-1 perform \textbf{-1.1\%/-3.1\%} respectively, which proves that our differentiate amplification and guiding attention to the dominant modality through confidence-based weights is crucial.


\subsection{Visualization Analysis}
\noindent\textbf{Multi-modal sample weight distribution.} 
We employ the T-SNE~\cite{tsne} dimensionality reduction technique to visualize and analyze the feature distributions of vehicles. In the Fig.~\ref{fig:tsne}, we visualize the feature distributions of module ablations compared with the baseline. Compared with Fig.~\ref{fig:tsne} (a) and (d), DCG further reduces the distance between instances of the same ID, making the features of each ID more compact and increasing the gap between different IDs. Compared with Fig.~\ref{fig:tsne} (b)(c) and (d), features for each ID become more compact and the gap between different IDs increases from our two fusion strategies. It fully demonstrates that our proposed method can expand intra-class consistency while reducing inter-modal differences, thereby achieving excellent discriminative ability.
\begin{figure}[htbp]  
\hfill  
\includegraphics[width=1.0\linewidth]{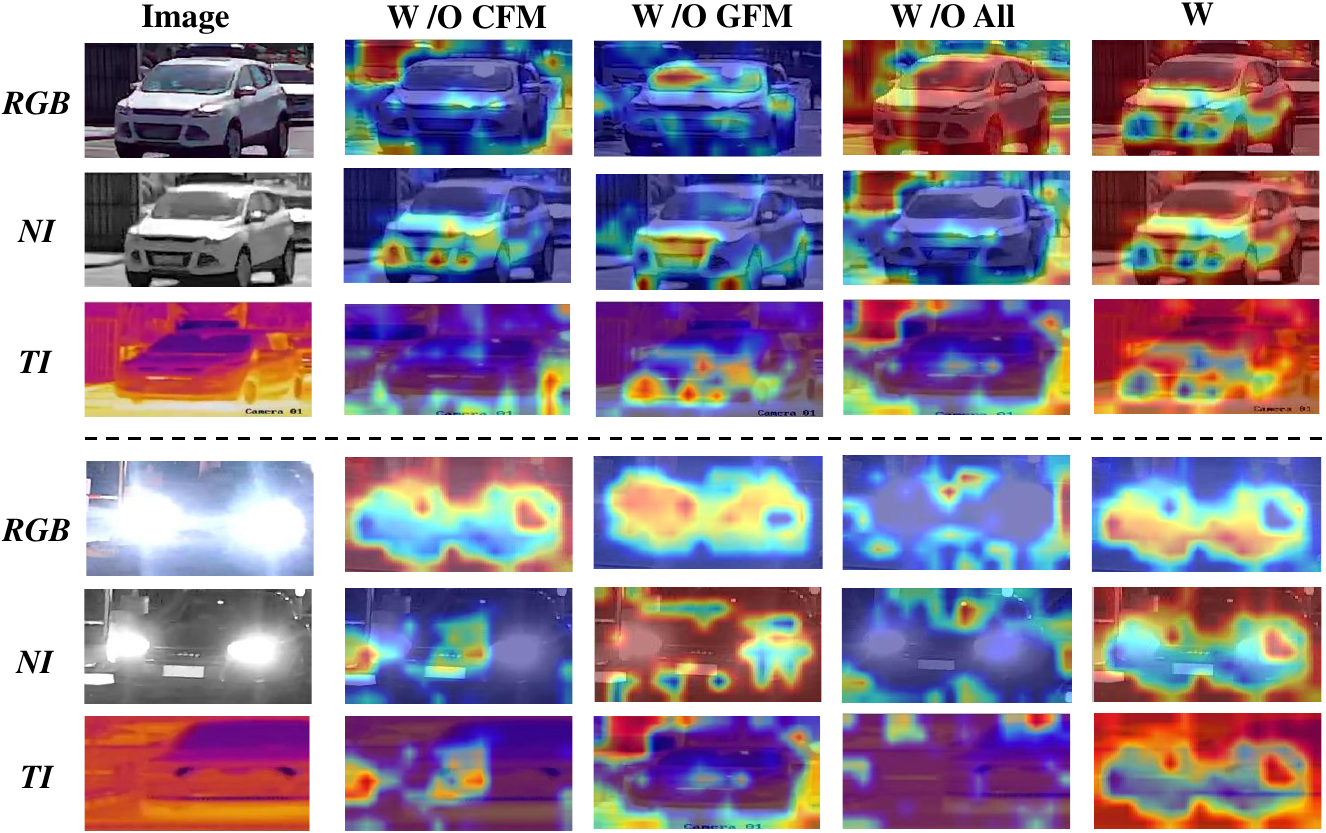}  
\caption{Class activation map visualization on WMVeID863
from different methods with two main challenges.}
\label{fig:cam}
\end{figure}
\begin{figure}[htbp]  
\hfill  
\includegraphics[width=1.0\linewidth]{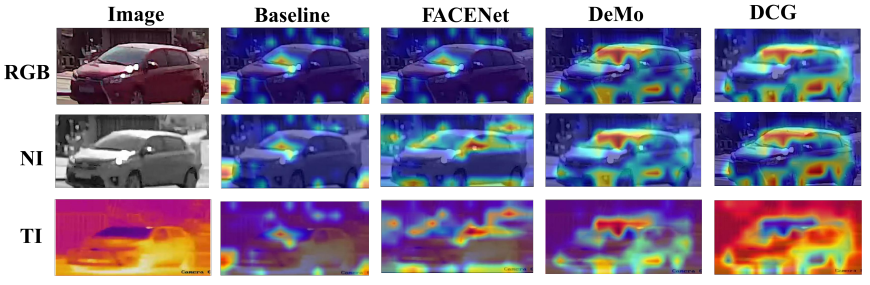}  
\caption{Class activation map visualization on
WMVeID863 comparing different methods.}
\label{fig:cam1}
\end{figure}

\noindent\textbf{Class Activation Map Visualization.}
To validate the effectiveness of the modules of our DCG, we present the Fig.~\ref{fig:cam} visualization of the ablation of the modules based on our method and the Fig.~\ref{fig:cam1} visualization of the comparison between our method and the current advanced methods. We present the channel activation maps of different methods on the WMVeID863~\cite{Facenet}. And to better show the effectiveness of our method dealing with the complex environment, we select the two main challenges faced by vehicles: low light and flare pollution to demonstrate the effectiveness of our fusion-decoupling framework in dealing with complex scenarios. Among Fig.~\ref{fig:cam}, ``w/o All" indicates all samples are fed into the two fusion strategies simultaneously, ``w/o CFM" and ``w/o GFM" denote our two fusion strategies respectively. As shown in Fig.~\ref{fig:cam}, introducing CFM and GFM significantly reduces feature response on background areas while excellently enhancing response on vehicles in the infrared-spectral modality. By comparing ``w/o All" and ``w", we can verify the effectiveness of our fusion decoupling method for avoiding fusion conflicts. Specifically, if multi-modal data is not fused separately based on confidence, comparative results show that it is impossible to focus on the vehicle itself effectively and accurately while avoiding interference from background noise.

Summarize the comparison of the above different methods which show the visualization diagrams at two levels, we can observe that in vehicle scenarios facing different challenges, our method can effectively focus on the vehicles themselves. This not only demonstrates the effectiveness of our module but also highlights the superiority of our fusion decoupling framework.
\section{Conclusion}
In this paper, we present a novel framework named DCG for multi-modal vehicle ReID. Our approach starts with a Dynamic Confidence-based Disentangling Weighting (DCDW) to differentiate multi-modal data with different quality distributions without fusion conflicts. Then, we introduce two fusion strategies named Collaboration Fusion Module (CFM) and Guidance Fusion Module (GFM) to achieve targeted fusion for uncertain modality quality distribution data. Among the two fusion methods, we respectively leverage adaptively and discard modalities with poor quality confidence-based weight adaptation to amplify the advantages of the dominant modality. We conduct experiments on three vehicle datasets WMVeID863, MSVR310 and RGBNT100 which with different challenges respectively and achieve state-of-the-art performance, which validates the effectiveness of our targeted fusion method. Moreover, to demonstrate the rationality of our fusion decoupling approach, we also conduct comparative experiments to prove that it outperforms the same fusion framework under the decoupling mode. In future work, we will further refine the potential of fusion conflicts and design more targeted fusion strategies for complex scenarios faced by vehicles to achieve greater gains.

\section{References Section}

\bibliographystyle{IEEEtran}  %
\bibliography{references}  %
\begin{IEEEbiography}
[{\includegraphics[width=1in,height=1.25in, clip,keepaspectratio]{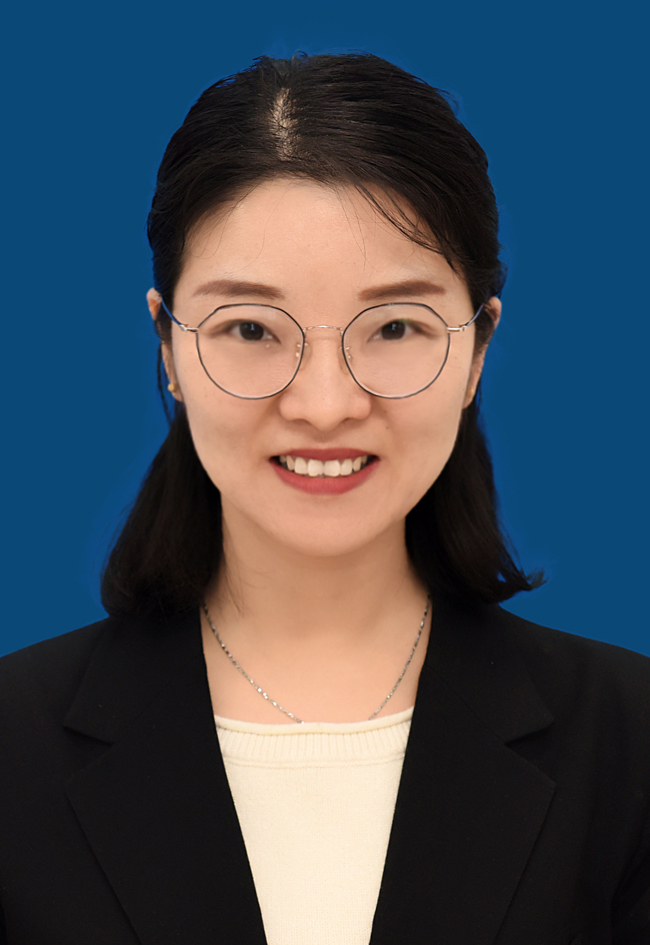}}]{Aihua Zheng} received the B.Eng. and the integrated master’s and Ph.D. degrees in computer science and technology from Anhui University, China, in 2006 and 2008, respectively, and the Ph.D. degree in computer science from the University of Greenwich, U.K., in 2012. She is currently a Professor of artificial intelligence at Anhui University. Her research interests include computer vision and artificial intelligence, especially on person/vehicle re-identification, audio-visual learning, and multi-modal and cross-modal learning.
\end{IEEEbiography}
\begin{IEEEbiography}
[{\includegraphics[width=1in,height=1.25in, clip,keepaspectratio]{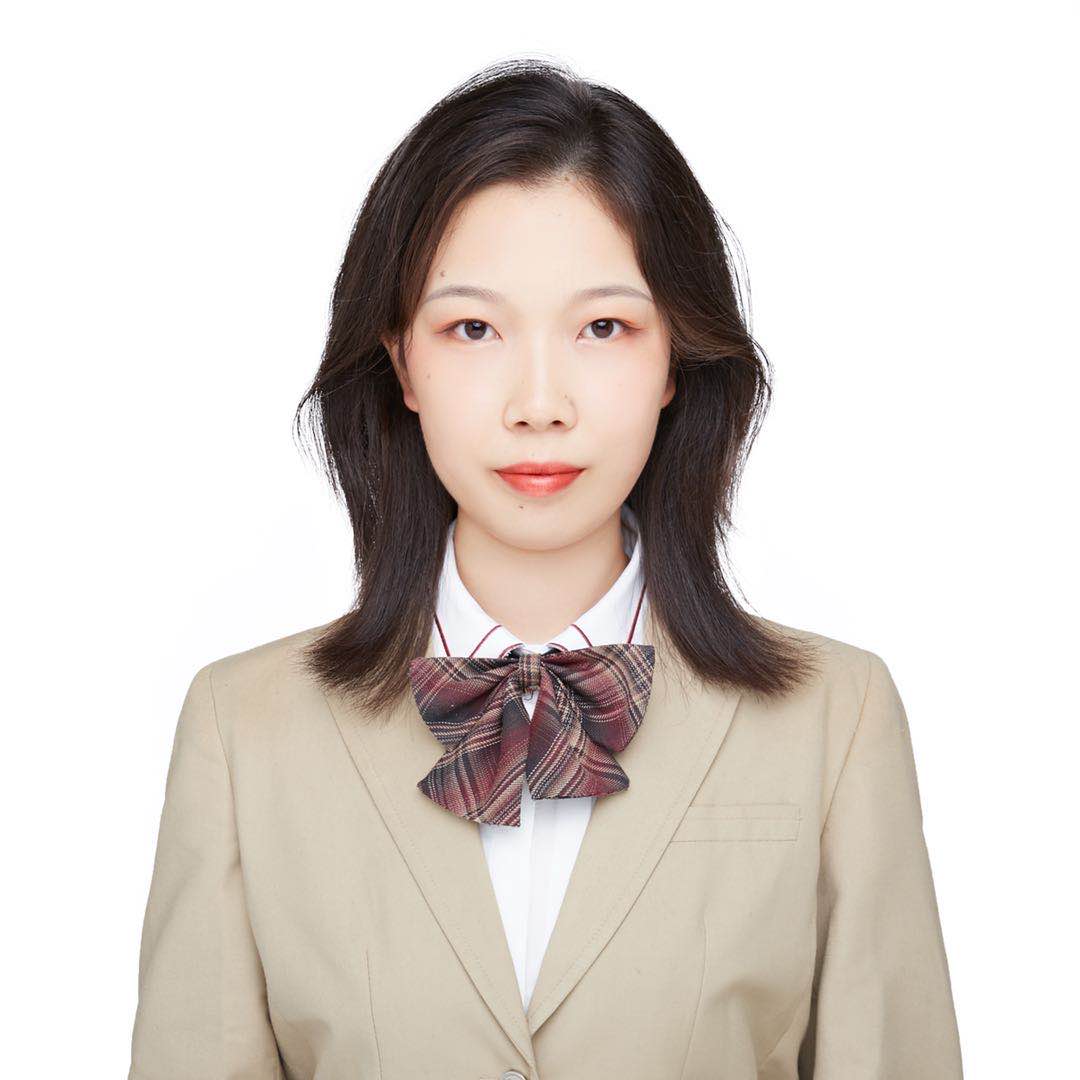}}] {Ya Gao} received her B.Eng. degree in 2024 and is currently pursuing the M.Eng degree in the School of Artificial Intelligence, Anhui University, Hefei, China. Her research interests include computer vision, multi-modal learning and vehicle
re-identification.
\end{IEEEbiography}
\vspace{-0.7\baselineskip} 
\begin{IEEEbiography}
[{\includegraphics[width=1in,height=1.25in, clip,keepaspectratio]{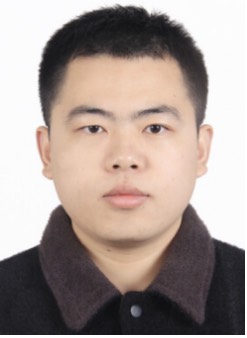}}] {Shihao Li} is currently pursuing the Ph.D. degree in Computer Science and Technology with the School of Artificial Intelligence, Anhui University. He received the B.Eng. degree from Tongling University, Tongling, China, in 2017. He is currently a Visiting Student at the Center for Research on Intelligent Perception and Computing (CRIPAC), National Laboratory of Pattern Recognition (NLPR), Institute of Automation, Chinese Academy of Sciences (CASIA), Beijing, China. His research interests include computer vision, multi-modal learning, and object re-identification.
\end{IEEEbiography}
\vspace{-0.7\baselineskip} 
\begin{IEEEbiography}
[{\includegraphics[width=1in,height=1.25in, clip,keepaspectratio]{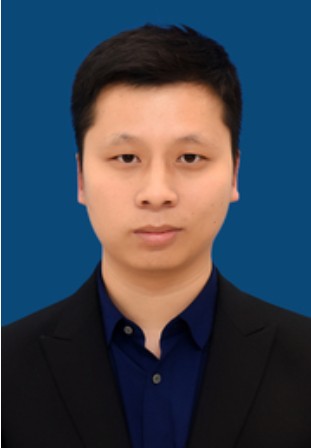}}] {Chenglong Li} received the M.S. and Ph.D. degrees from the School of Computer Science and Technology, Anhui University, Hefei, China, in 2013 and 2016, respectively. From 2014 to 2015, he was a Visiting Student at the School of Data and Computer Science, Sun Yat-sen University, Guangzhou, China. He was a Postdoctoral Research Fellow with the Center for Research on Intelligent Perception and Computing (CRIPAC), National Laboratory of Pattern Recognition (NLPR), Institute of Automation, Chinese Academy of Sciences (CASIA), Beijing, China. He is currently a Professor of artificial intelligence at Anhui University. His current research interests include computer vision and deep learning. He was a recipient of the ACM Hefei Doctoral Dissertation Award in 2016.
\end{IEEEbiography}
\vspace{-0.7\baselineskip} 
\begin{IEEEbiography}
[{\includegraphics[width=1in,height=1.25in, clip,keepaspectratio]{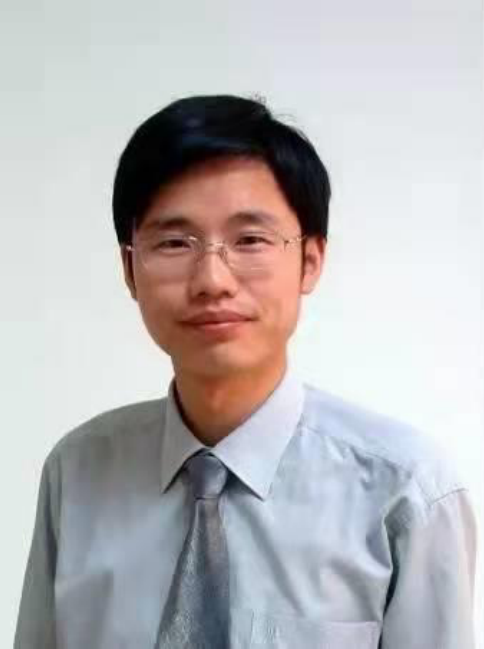}}] {Jin Tang} received the B.Eng. degree in automation and the Ph.D. degree in Computer Science from
Anhui University, Hefei, China, in 1999 and 2007, respectively. He is currently a Professor and PhD supervisor with the School of Computer Science and Technology, Anhui University. His research interests include computer vision, pattern recognition, and machine learning.
\end{IEEEbiography}
\end{document}